\documentclass[11pt, a4paper]{lumia}
\usepackage[sort&compress]{natbib}
\setheadertext{FlowRL: Matching Reward Distributions for LLM Reasoning}

\renewcommand{\today}{2025-09-17}

\usepackage{xspace}
\newcommand*{\eg}{\textit{e.g.}\xspace}

\theoremstyle{plain}
\newtheorem{theorem}{Theorem}%[section]
\newtheorem{proposition}[theorem]{Proposition}
\newtheorem*{proposition*}{Proposition}

\theoremstyle{definition}

\theoremstyle{definition}
\newtheorem{remark}[theorem]{Remark}

\def\eqref#1{equation~\ref{#1}}
\usepackage[utf8]{inputenc}     % allow utf-8 input
\usepackage[T1]{fontenc}        % use 8-bit T1 fonts
\usepackage{microtype}          % microtypography

\usepackage{xcolor}             % colors
\usepackage[dvipsnames]{xcolor} % extended color names
\usepackage{graphicx}           % graphics support
\usepackage{tikz}               % TikZ graphics
\usepackage[edges]{forest}      % forest trees

\usepackage{hyperref}           % hyperlinks
\usepackage{url}                % simple URL typesetting
\usepackage{xurl}               % extended URL breaking

\usepackage{tabularx}           % extended tabular
\usepackage{array}              % extended array and tabular
\usepackage{booktabs}           % professional tables
\usepackage{longtable}          % multi-page tables
\usepackage{multirow}           % multirow cells
\usepackage{makecell}           % enhanced cell formatting
\usepackage[table]{xcolor}      % colored tables
\usepackage{colortbl}           % colored table cells
\usepackage{ragged2e}           % ragged text alignment

\usepackage{amsmath}            % enhanced math
\usepackage{mathtools}          % additional math tools
\usepackage{amsfonts}           % blackboard math symbols
\usepackage{amssymb}            % additional math symbols
\usepackage{amsthm}             % theorem environments
\usepackage{nicefrac}           % compact symbols for 1/2, etc.

\usepackage{algorithm}          % algorithm environment
\usepackage{algorithmicx}       % algorithmic pseudocode
\usepackage{algpseudocode}      % pseudocode formatting
\usepackage{listings}           % code listings

\usepackage{subcaption}         % subfigures and subcaptions
\usepackage{caption}            % enhanced captions
\usepackage{wrapfig}            % text wrapping around figures
\usepackage[export]{adjustbox}  % adjustable boxes

\usepackage{soul}               % text highlighting and underlining
\usepackage{changes}            % track changes
\usepackage{xspace}             % intelligent spacing
\usepackage[normalem]{ulem}     % underlining without affecting emphasis
\usepackage{CJKutf8}            % CJK character support

\usepackage{enumitem}           % enhanced lists
\usepackage{pifont}             % special symbols (including ding)

\usepackage[most,skins,theorems]{tcolorbox} % colored boxes
\usepackage[tikz]{bclogo}       % logo boxes
\usepackage[framemethod=tikz]{mdframed} % framed text

\usepackage{lipsum}             % lorem ipsum text
\usepackage{tocloft}            % table of contents formatting
\usepackage{fontawesome5}       % FontAwesome icons
\usepackage{afterpage}          % page break control
\usepackage{bbding}             % additional symbols
\usepackage{epigraph}           % epigraphs
\usepackage{minitoc}            % mini table of contents
\usepackage{multicol}           % multiple columns
\usepackage{textgreek}          % Greek text

\newcolumntype{P}[1]{>{\RaggedRight\arraybackslash}p{#1}}

\definecolor{darkblue}{rgb}{0, 0, 0.5}
\definecolor{uclablue}{RGB}{39, 116, 174}
\definecolor{bigaired}{RGB}{156, 0, 0}
\definecolor{myblue}{HTML}{598BE7}
\definecolor{mildblue}{RGB}{31,119,180}
\definecolor{sectionblue}{RGB}{70, 130, 180}
\definecolor{methodblue}{RGB}{0, 150, 136}
\definecolor{bgblue}{RGB}{245,243,253}
\definecolor{ttblue}{RGB}{91,194,224}
\definecolor{mygreen}{rgb}{0.64, 0.56, 0.88}
\definecolor{myyellow}{rgb}{0.68, 0.6, 0.1}
\definecolor{fancygreen}{rgb}{0.33, 0.68, 0.20}
\definecolor{salmon}{rgb}{0.94, 0.52, 0.49}
\definecolor{tablegreen}{rgb}{0.82, 0.94, 0.75}
\definecolor{tableblue}{rgb}{0.81, 0.90, 0.94}
\definecolor{tablered}{rgb}{0.97, 0.85, 0.85}
\definecolor{tableorange}{rgb}{0.96, 0.85, 0.81}
\definecolor{myorange}{rgb}{1.0, 0.49, 0.0}
\definecolor{tlgreen}{rgb}{0.33, 0.68, 0.20}
\definecolor{darkgreen}{RGB}{0,100,0}
\definecolor{darkred}{RGB}{200, 0, 0}
\definecolor{lightblue}{RGB}{220,235,250}
\definecolor{customyellow}{HTML}{FFFACD}
\definecolor{refinegreen}{RGB}{0, 128, 75}
\definecolor{scoregreen}{RGB}{34, 139, 34}
\definecolor{hidden-blue}{RGB}{194,232,247}
\definecolor{hidden-black}{RGB}{20,68,106}
\definecolor{yes}{HTML}{C6EFCE}
\definecolor{no}{HTML}{FFC7CE}
\definecolor{partial}{HTML}{FFEB9C}
\definecolor{external}{HTML}{D9E1F2}
\definecolor{hdr}{HTML}{F2F2F2}
\definecolor{GRPOrow}{gray}{0.96}
\definecolor{FlowRLrow}{RGB}{225,236,255}
\definecolor{FlowBlue}{RGB}{80,120,210}
\definecolor{GRPOGray}{gray}{0.35}

\hypersetup{
    colorlinks=true, 
    citecolor=uclablue, 
    linkcolor=bigaired,
    urlcolor=darkblue
}

\setlist[itemize]{leftmargin=20pt, noitemsep, topsep=0pt}

\NewDocumentCommand{\kaiyan}{mO{}}{\textcolor{purple}{\textsuperscript{\textit{kaiyan}}\textsf{\textbf{\small[#1]}}}}
\NewDocumentCommand{\yuxin}{mO{}}{\textcolor{cyan}{\textsuperscript{\textit{yuxin}}\textsf{\textbf{\small[#1]}}}}
\NewDocumentCommand{\bx}{mO{}}{\textcolor{green}{\textsuperscript{\textit{bx}}\textsf{\textbf{\small[#1]}}}}
\NewDocumentCommand{\at}{mO{}}{\textcolor{red}{\textsuperscript{\textit{AT}}\textsf{\textbf{\small[#1]}}}}
\NewDocumentCommand{\re}{mO{}}{\textcolor{blue}{\textsuperscript{\textit{RE}}\textsf{\textbf{\small[#1]}}}}
\NewDocumentCommand{\ybsun}{mO{}}{\textcolor{magenta}{\textsuperscript{\textit{youbang}}\textsf{\textbf{\small[#1]}}}}
\NewDocumentCommand{\runze}{mO{}}{\textcolor{orange}{\textsuperscript{\textit{runze}}\textsf{\textbf{\small[#1]}}}}
\NewDocumentCommand{\add}{mO{}}{\textcolor{darkgreen}{\textsuperscript{\textit{Maybe Consider Discuss}}\textsf{\textbf{[#1]}}}}

\newcommand{\cmark}{\textcolor{darkgreen}{\boldmath$\checkmark$}}
\newcommand{\xmark}{\textcolor{darkred}{\boldmath$\times$}}

\newcommand{\gact}[1]{\fcolorbox{black!40}{gray!12}{\strut #1}}
\newcommand{\fact}[1]{\fcolorbox{FlowBlue!60!black}{FlowRLrow}{\strut #1}}
\newcommand{\gmini}[1]{\fcolorbox{black!30}{gray!06}{\scriptsize #1}}
\newcommand{\repbadgeR}[1]{\,\raisebox{.25ex}{\scriptsize\colorbox{red!10}{\(\times\)\textcolor{red!75!black}{#1}}}}

\DeclareMathOperator*{\argmax}{arg\,max}
\DeclareMathOperator*{\argmin}{arg\,min}

\newenvironment{itemize*}%
 {\leftmargini=10pt\begin{itemize}%
  \setlength{\itemsep}{0pt}%
  \setlength{\parskip}{0pt}%
  }%
 {\end{itemize}}

\newenvironment{enumerate*}%
 {\begin{enumerate}%
  \setlength{\itemsep}{0pt}%
  \setlength{\parskip}{0pt}}%
 {\end{enumerate}}

\newcommand{\cellstatus}[1]{%
  \begingroup
  \StrTrim{#1}[\statusval]%
  \IfStrEq{\statusval}{Yes}{\cellcolor{yes}\cmark}{}%
  \IfStrEq{\statusval}{No}{\cellcolor{no}\xmark}{}%
  \IfBeginWith{\statusval}{Yes (}{\cellcolor{yes}\cmark~\textit{\statusval\unskip}}{}%
  \IfStrEq{\statusval}{Partial}{\cellcolor{partial}\textbf{Partial}}{}%
  \IfStrEq{\statusval}{External}{\cellcolor{external}\textbf{External}}{}%
  \endgroup
}

\newtcolorbox{myboxi}[1][]{
  breakable,
  title=#1,
  colback=red!5,
  colbacktitle=red!5,
  coltitle=black,
  fonttitle=\bfseries,
  bottomrule=0pt,
  toprule=0pt,
  leftrule=2pt,
  rightrule=2pt,
  titlerule=0pt,
  arc=0pt,
  outer arc=0pt,
  colframe=red,
}

\newtcolorbox{myboxnote}[1][]{
  breakable,
  title=#1,
  colback=orange!0,
  colbacktitle=orange!0,
  coltitle=black,
  fonttitle=\bfseries,
  bottomrule=0pt,
  toprule=0pt,
  leftrule=2pt,
  rightrule=2pt,
  titlerule=0pt,
  arc=0pt,
  outer arc=0pt,
  colframe=orange,
}

\newtcolorbox{myboxii}[1][]{
  breakable,
  freelance,
  title=#1,
  colback=white,
  colbacktitle=white,
  coltitle=black,
  fonttitle=\bfseries,
  bottomrule=0pt,
  boxrule=0pt,
  colframe=white,
  overlay unbroken and first={
  \draw[red!75!black,line width=3pt]
    ([xshift=5pt]frame.north west) -- 
    (frame.north west) -- 
    (frame.south west);
  \draw[red!75!black,line width=3pt]
    ([xshift=-5pt]frame.north east) -- 
    (frame.north east) -- 
    (frame.south east);
  },
  overlay unbroken app={
  \draw[red!75!black,line width=3pt,line cap=rect]
    (frame.south west) -- 
    ([xshift=5pt]frame.south west);
  \draw[red!75!black,line width=3pt,line cap=rect]
    (frame.south east) -- 
    ([xshift=-5pt]frame.south east);
  },
  overlay middle and last={
  \draw[red!75!black,line width=3pt]
    (frame.north west) -- 
    (frame.south west);
  \draw[red!75!black,line width=3pt]
    (frame.north east) -- 
    (frame.south east);
  },
  overlay last app={
  \draw[red!75!black,line width=3pt,line cap=rect]
    (frame.south west) --
    ([xshift=5pt]frame.south west);
  \draw[red!75!black,line width=3pt,line cap=rect]
    (frame.south east) --
    ([xshift=-5pt]frame.south east);
  },
}

\tcbset{
  takeawaysbox/.style={
    title=Takeaways,
    colback=lightblue!80,
    colframe=black,
    fonttitle=\bfseries\small,
    coltitle=white,
    colbacktitle=black,
    enhanced,
    attach boxed title to top left={xshift=2.5mm,yshift=-2.5mm},
    boxed title style={rounded corners, size=small, colframe=black, colback=black},
    width=\linewidth,
    arc=3.5mm
  }
}

\mdfdefinestyle{mystyle}{%
  rightline=true,
  innerleftmargin=10,
  innerrightmargin=10,
  outerlinewidth=3pt,
  topline=false,
  rightline=true,
  bottomline=false,
  skipabove=\topsep,
  skipbelow=\topsep
}

\tikzset{%
    every node/.style={font=\tiny},
    parent/.style =          {align=center,text width=2cm,rounded corners=3pt, line width=0.3mm, fill=gray!10,draw=gray!80},
    child/.style =           {align=center,text width=2.0cm,rounded corners=3pt, fill=blue!10,draw=blue!80,line width=0.3mm},
    grandchild/.style =      {align=center,text width=2cm,rounded corners=3pt},
    greatgrandchild/.style = {align=center,text width=1.5cm,rounded corners=3pt},
    greatgrandchild2/.style = {align=center,text width=1.5cm,rounded corners=3pt},    
    referenceblock/.style =  {align=center,text width=1.5cm,rounded corners=2pt},
    pretrain/.style =           {align=center,text width=2.0cm,rounded corners=3pt, fill=blue!10,draw=blue!80,line width=0.3mm},   
    pretrain_work/.style =           {align=center, text width=8.5cm,rounded corners=3pt, fill=blue!10,draw=blue!0,line width=0.3mm},  
    template/.style =           {align=center,text width=2.0cm,rounded corners=3pt, fill=red!10,draw=red!80,line width=0.3mm},   
    template_work/.style =           {align=center,text width=8.5cm,rounded corners=3pt, fill=red!10,draw=red!0,line width=0.3mm},    
    answer/.style =           {align=center,text width=2.0cm,rounded corners=3pt, fill= cyan!10,draw= cyan!80,line width=0.3mm},   
    answer_work/.style =           {align=center,text width=8.5cm,rounded corners=3pt, fill= cyan!10,draw= cyan!0,line width=0.3mm},      
    multiple/.style =           {align=center,text width=2.0cm,rounded corners=3pt, fill= orange!10,draw= orange!80,line width=0.3mm},   
    multiple_work/.style =           {align=center,text width=8.5cm,rounded corners=3pt, fill= orange!10,draw= orange!0,line width=0.3mm},        
    tuning/.style =           {align=center,text width=2.0cm,rounded corners=3pt, fill= magenta!10,draw= magenta!80,line width=0.3mm},   
    tuning_work/.style =           {align=center,text width=8.5cm,rounded corners=3pt, fill= magenta!10,draw= magenta!0,line width=0.3mm},          
}

\newcommand{\lstbg}[3][0pt]{{\fboxsep#1\colorbox{#2}{\strut #3}}}

\lstdefinelanguage{diff}{
  basicstyle=\ttfamily\small,
  morecomment=[f][\lstbg{red!20}]-,
  morecomment=[f][\lstbg{green!20}]+,
}

\lstdefinelanguage{diffpython}{
  language=diff,
  morekeywords={def, if, else, for, while, return, import, from, as, class, with, try, except, finally, raise, lambda, and, or, not, in, is, None, True, False},
  morecomment=[l]{\#},
  morestring=[b]",
  morestring=[b]',
}

\title{FlowRL: Matching Reward Distributions for LLM Reasoning}

\author{%
    \scriptsize Xuekai Zhu$^{1}$, Daixuan Cheng$^{6}$, Dinghuai Zhang$^{3}$, Hengli Li$^{5}$, Kaiyan Zhang$^{4}$, Che Jiang$^{4}$,  Youbang Sun$^{4}$, Ermo Hua$^{4}$, Yuxin Zuo$^{4}$, Xingtai Lv$^{4}$, Qizheng Zhang$^{7}$, Lin Chen$^{1}$, Fanghao Shao$^{1}$, Bo Xue$^{1}$, Yunchong Song$^{1}$, Zhenjie Yang$^{1}$, Ganqu Cui$^{2}$, Ning Ding$^{4,2}$, Jianfeng Gao$^{3}$, Xiaodong Liu$^{3}$, 
    Bowen Zhou$^{4,2\ddagger}$, Hongyuan Mei$^{8\ddagger}$, Zhouhan Lin$^{1,2\ddagger}$ \\
    $^1$ Shanghai Jiao Tong University \quad
    $^2$ Shanghai AI Laboratory \quad
    $^3$ Microsoft Research \quad
    $^4$ Tsinghua University \quad
    $^5$ Peking University \quad
    $^6$ Renmin University of China \quad
    $^7$ Stanford University \quad
    $^8$ Toyota Technological Institute at Chicago \\
    \faEnvelope[regular]~\texttt{hongyuanmei@gmail.com}  \quad
    \faEnvelope[regular]~\texttt{xuekaizhu0@gmail.com}  \quad
    \faGithub~\href{https://github.com/Xuekai-Zhu/FlowRL}{FlowRL} \quad $^\ddagger$ Corresponding Authors. 
}

\begin{abstract}
We propose FlowRL: matching the full reward distribution via flow balancing instead of maximizing rewards in large language model (LLM) reinforcement learning (RL).
Recent advanced reasoning models adopt reward-maximizing methods (\eg, PPO and GRPO), which tend to over-optimize dominant reward signals while neglecting less frequent but valid reasoning paths, thus reducing diversity.
In contrast, we transform scalar rewards into a normalized target distribution using a learnable partition function, and then minimize the reverse KL divergence between the policy and the target distribution.
We implement this idea as a flow-balanced optimization method that promotes diverse exploration and generalizable reasoning trajectories.
We conduct experiments on math and code reasoning tasks: FlowRL achieves a significant average improvement of $10.0\%$ over GRPO and $5.1\%$ over PPO on math benchmarks, and performs consistently better on code reasoning tasks. These results highlight reward distribution-matching as a key step toward efficient exploration and diverse reasoning in LLM reinforcement learning.
\end{abstract}

\begin{document}
\maketitle
\vspace{-1em}
\begin{figure}[h]
    \centering
    \includegraphics[width=0.95\textwidth]{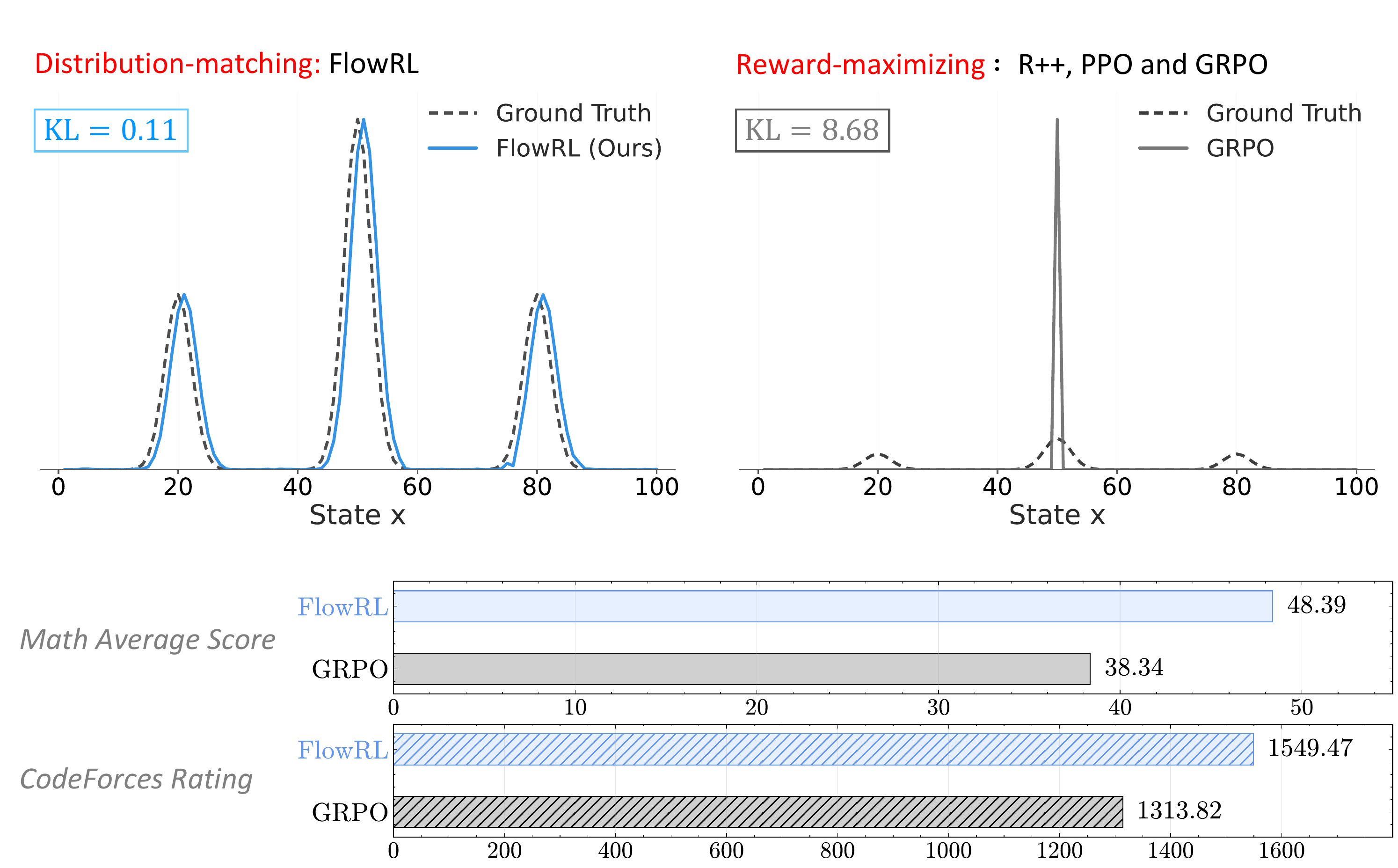}
    \vspace{-10pt}
    \caption{\textbf{Top}: Comparison between distribution-matching and reward-maximizing approaches. FlowRL (left) learns to match the full reward distribution, maintaining diversity across multiple modes with low KL divergence. In contrast, reward-maximizing methods like GRPO (right) concentrate on a single high-reward peak, leading to mode collapse and higher KL divergence.
    \textbf{Bottom}: Performance comparison. FlowRL consistently outperforms GRPO across math and code domains.}
    \label{fig:main_fig}
\end{figure}
% Detailed results see in~\S~\ref{sec:results}.
%\vspace{-10pt}

\section{Introduction}
Reinforcement learning (RL) plays a crucial role in the post-training of large language models (LLMs)~\citep{zhang2025survey}. A series of powerful reasoning models~\citep{guo2025deepseek,Gemini25,rastogi2025magistral} have employed large-scale reinforcement learning to achieve strong performance on highly challenging benchmarks~\citep{he2024olympiadbench}.
The evolution of RL algorithms for LLM reasoning has progressed through several key stages: REINFORCE~\citep{sutton1999reinforcement} provides a solid baseline that is easy to implement and efficient in simple settings; PPO~\citep{schulman2017proximal} improves upon REINFORCE with better stability and efficiency in complex settings; GRPO~\citep{shao2024deepseekmath} simplifies PPO training by eliminating value functions and relying on group comparisons, though at the cost of requiring more rollouts per update. However, all these methods share a fundamental limitation in their reward-maximizing objective.

Reward-maximizing RL methods tend to overfit to the dominant mode of the reward distribution~\citep{skalse2022defining,pan2022effects,zelikman2022star,gao2023scaling}. 
This often results in limited diversity among generated reasoning paths and reduces generalization to less frequent yet valid logical outcomes~\citep{hu2023amortizing}. As illustrated in Figure~\ref{fig:main_fig}, GRPO neglects other meaningful modes. These drawbacks become especially pronounced in complex long chain-of-thought~(CoT;~\citealp{wei2022chain}) reasoning, where capturing a diverse distribution of plausible solutions is essential for effective generalization~\citep{liu2025scaling}. Recent approaches adjust the clip ratio~\citep{yu2025dapo}, augment the advantage function with an entropy-based term~\citep{cheng2025reasoning}, or selectively promote high-entropy tokens~\citep{wang2025beyond}, thereby dynamically adapting the training data distribution and implicitly increasing diversity during training. This raises a fundamental question: How can we promote diverse exploration to prevent convergence to dominant solution patterns in RL training?

In this paper, we propose \textbf{FlowRL}, a policy optimization algorithm that aligns the policy model with the full reward distribution, encouraging mode coverage. 
FlowRL achieves more efficient exploration by fundamentally shifting from reward maximization to reward distribution matching, thereby addressing the inherent mode-collapse limitations of previous RL approaches. As illustrated in Figure~\ref{fig:main_fig}, the core idea of FlowRL is to introduce a learnable partition function that normalizes scalar rewards into a target distribution, and to minimize the reverse KL divergence between the policy and this reward-induced distribution.
We develop this KL objective based on the trajectory balance formulation from GFlowNets~\citep{bengio2023gflownet}, providing a gradient equivalence proof that bridges generative modeling and policy optimization. To address the challenges of long CoT training, we introduce two key technical solutions: \textit{length normalization} to tackle gradient explosion issues that occur with variable-length CoT reasoning, and \textit{importance sampling} to correct for the distribution mismatch between generated rollouts and the current policy.

We compare FlowRL with mainstream RL algorithms including REINFORCE++, PPO, and GRPO across math and code domains, using both base and distilled LLMs (7B, 32B). In math domain, FlowRL outperforms GRPO and PPO by $10.0\%$ and $5.1\%$, respectively, demonstrating consistent improvements across six challenging math benchmarks. Furthermore, FlowRL surpasses both PPO and GRPO on three challenging coding benchmarks, highlighting its strong generalization capabilities in code reasoning tasks. To understand what drives these performance gains, we analyze the diversity of generated reasoning paths. This diversity analysis confirms that FlowRL generates substantially more diverse rollouts than baseline methods, validating our approach's effectiveness in exploring multiple solution strategies. 
%Case studies further demonstrate how FlowRL's diverse exploration finds successful solutions where GRPO gets trapped in repetitive loops.

\paragraph{Contributions.} We summarize the key contributions of this
work as follows:
\begin{itemize}[leftmargin=*, topsep=0pt, noitemsep]
    \item We propose FlowRL, a policy optimization algorithm that shifts from reward maximization to reward distribution matching via flow balance, encouraging diverse reasoning path exploration while addressing the inherent mode-collapse limitations of existing RL methods.
    \item We introduce length normalization and importance sampling to enable effective training on variable-length CoT reasoning, addressing gradient explosion and sampling mismatch issues.
    \item FlowRL outperforms GRPO and PPO by 10.0\% and 5.1\% respectively across math benchmarks and demonstrates strong generalization on code reasoning tasks, with diversity analysis confirming substantially more diverse solution exploration.
\end{itemize}

\section{Preliminaries}\label{sec:preliminaries}
\paragraph{Reinforcement Learning for Reasoning.}
We formulate reasoning as a conditional generation problem, where the policy model receives a question $\mathbf{x} \in \mathcal{X}$ and generates an answer $\mathbf{y} \in \mathcal{Y}$. The objective is to learn a policy $\pi_\theta(\mathbf{y}|\mathbf{x})$ that produces high-quality answers under task-specific reward signals $r$. 
To better illustrate the policy optimization procedure, we provide a detailed formulation of GRPO below. For each question $\mathbf{x}$, GRPO samples a group of answers $\{\mathbf{y}_1, \mathbf{y}_2, \dots, \mathbf{y}_G\}$ from old policy $\pi_{\theta_{old}}$ and updates the model by maximizing the following objective:
\begin{equation}
\resizebox{\linewidth}{!}{$
\begin{split}
    \mathcal{J}_{GRPO}(\theta) &= \mathbb{E}_{[\mathbf{x} \sim P(\mathcal{X}), \{\mathbf{y}_i\}_{i=1}^G \sim \pi_{\theta_{old}}(\mathcal{Y}|\mathbf{x})]}  \\
    & \frac{1}{G}\sum_{i=1}^G\frac{1}{|\mathbf{y}_i|} \sum_{t=1}^{|\mathbf{y}_i|} \left\{ \min \left[ \frac{\pi_\theta(\mathbf{y}_{i,t} | \mathbf{x}, \mathbf{y}_{i,<t})}{\pi_{\theta_{old}}(\mathbf{y}_{i,t} | \mathbf{x}, \mathbf{y}_{i,<t})} \hat{A}_{i,t}, \text{clip} \left( \frac{\pi_\theta(\mathbf{y}_{i,t} | \mathbf{x}, \mathbf{y}_{i,<t})}{\pi_{\theta_{old}}(\mathbf{y}_{i,t} | \mathbf{x}, \mathbf{y}_{i,<t})}, 1 - \epsilon, 1 + \epsilon \right)  \hat{A}_{i,t} \right] -  \lambda\mathbb{D}_{KL}\left[\pi_{\theta} || \pi_{ref}\right]\right\} , \\
    & \mathbb{D}_{\mathrm{KL}}(\pi_\theta \| \pi_{\mathrm{ref}}) = \frac{\pi_{\mathrm{ref}}(\mathbf{y}_i|\mathbf{x})}{\pi_\theta(\mathbf{y}_i|\mathbf{x})} - \log \frac{\pi_{\mathrm{ref}}(\mathbf{y}_i|\mathbf{x})}{\pi_\theta(\mathbf{y}_i|\mathbf{x})} - 1, 
\end{split}\label{eq:GRPO-obj}
$}
\end{equation}
where $\epsilon$ and $\lambda$ are hyper-parameters. Here, $A_i$ denotes the advantage, computed by normalizing the group reward values $\{r_1, r_2, \dots, r_G\}$ as $A_i = \frac{r_i - \mathrm{mean}(\{r_1, r_2, \cdots, r_G\})}
{\mathrm{std}(\{r_1, r_2, \cdots, r_G\})}.$
Compared to GRPO, REINFORCE applies the policy gradient directly, without advantage normalization, clipping, or KL regularization. PPO uses a critic model to estimate the advantage and employs importance sampling to stabilize policy updates.

\paragraph{GFlowNets.}
\begin{wrapfigure}{r}{0.5\textwidth}
    \vspace{-3em}
    \centering
    \begin{center}
        \includegraphics[width=\linewidth]{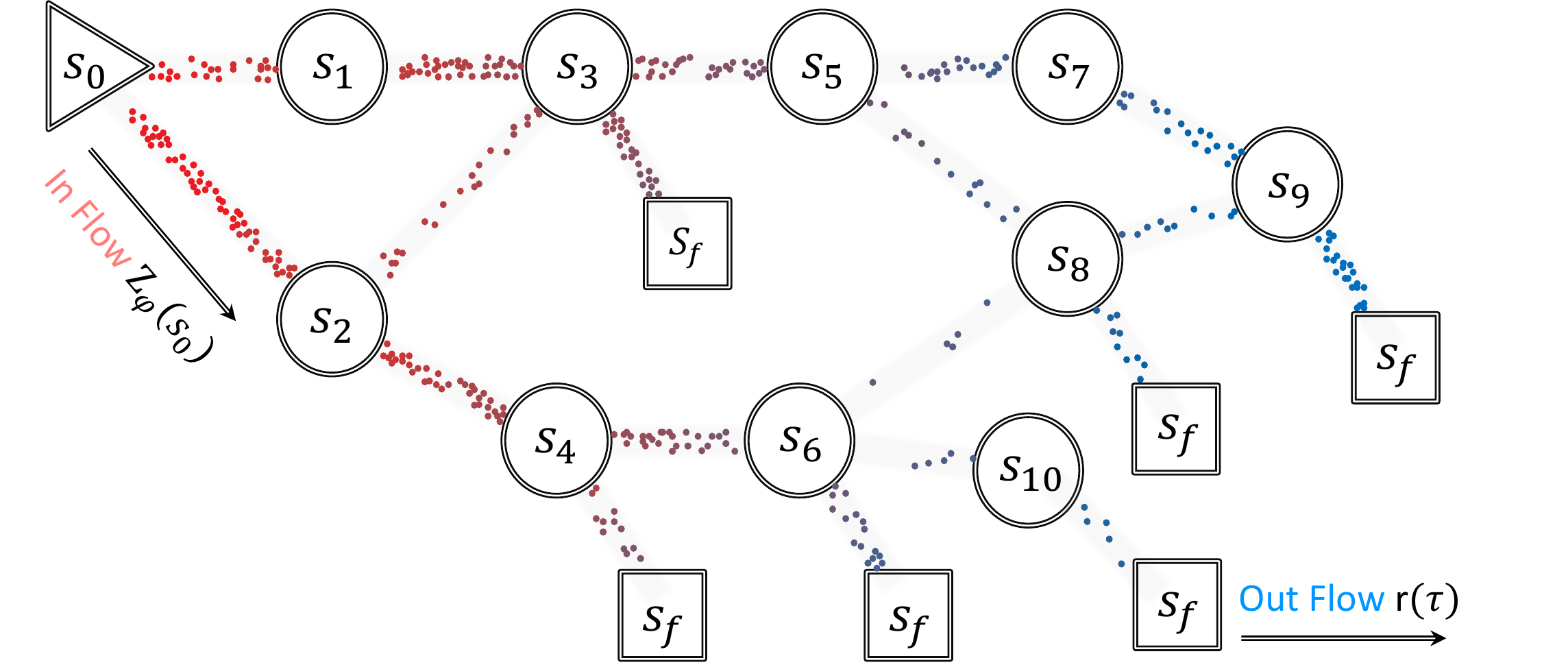}        
        \vspace{-2em}
        \caption{GFlowNets~\citep{JMLR:v24:22-0364}, a flow-balance perspective on reinforcement learning. 
     The initial flow $Z_\phi(s_0)$ injects probability mass into the environment, which is transported through intermediate states by the policy $\pi_\theta$ and accumulated at terminal states in proportion to the scalar rewards.}\label{fig:gflownet}
    \end{center}
    \vspace{-1em}
\end{wrapfigure}
Generative Flow Networks ~\citep{JMLR:v24:22-0364} are a probabilistic framework for training stochastic policies to sample discrete, compositional objects (\eg, graphs, sequences) in proportion to a given reward. As shown in Figure~\ref{fig:gflownet}, the core principle of GFlowNets is to balance the forward and backward probability flows at each state, inspired by flow matching~\citep{bengio2021flow}. The initial flow is estimated by $Z_\phi(s_0)$ at the initial state $s_0$. The output flow is equal to the outcome reward $r(s_n)$ conditioned at the final state $s_n$. Following~\cite{lee2024learning}, we use a 3-layer MLP to parameterize $Z_\phi$.
This flow-balancing mechanism facilitates the discovery of diverse, high-reward solutions by ensuring proper exploration of the solution space.  See Appendix~\ref{sec:gflownets_notation} for detailed GFlowNets background.

\section{Methodology}\label{sec:method}
In this section, we first formulate distribution matching in reinforcement learning through reverse KL divergence and establish its connection to trajectory balance from GFlowNets. To address the challenges of gradient explosion and sampling mismatch encountered during long CoT training, we further incorporate length normalization and importance sampling. Using this enhanced framework, we derive a flow-balanced objective, termed \textit{FlowRL}.

\subsection{From Reward Maximization to Distribution Matching}
As illustrated in Figure~\ref{fig:main_fig}, recent powerful large reasoning models typically employ reward-maximizing RL algorithms, such as PPO or GRPO. However, these methods tend to optimize toward the dominant reward mode, frequently resulting in mode collapse and the neglect of other plausible, high-quality reasoning paths. To address this fundamental limitation, we propose optimizing the policy by aligning its output distribution to a target reward distribution. A simple yet effective way to achieve this is to minimize the reverse KL divergence\footnote{We use reverse KL since we can only sample from the policy model, not the target reward distribution.} between the policy and this target. However, in long CoT reasoning tasks, the available supervision in RL is a scalar reward, rather than a full distribution. Moreover, enumerating or sampling all valid trajectories to recover the true reward distribution is computationally intractable.

Inspired by energy-based modeling~\citep{Hinton1995TheA, du2019implicit},
% arbel2020generalized
we introduce a learnable partition function $Z_\phi(\mathbf{x})$ to normalize scalar rewards into a valid target distribution. This allows us to minimize the reverse KL divergence between the policy and the reward-weighted distribution, formalized as:
\begin{equation}
\min_\theta \mathcal{D}_{\mathrm{KL}} \left( 
\pi_\theta(\mathbf{y} \mid \mathbf{x}) \,\middle\|\, 
\frac{\exp(\beta r(\mathbf{x}, \mathbf{y}))}{Z_\phi(\mathbf{x})}
\right) 
\quad \Rightarrow \quad 
\pi_\theta(\mathbf{y} \mid \mathbf{x}) \propto \exp(\beta r(\mathbf{x}, \mathbf{y})),
\label{eq:kl_objective}
\end{equation}
where $r(\mathbf{x}, \mathbf{y})$ is the reward function, $\beta$ is a hyperparameter, $Z_\phi(\mathbf{x})$ is the learned partition function, and the resulting target distribution is defined as $\tilde{\pi}(\mathbf{y} \mid \mathbf{x}) = \frac{\exp(\beta r(\mathbf{x}, \mathbf{y}))}{Z_\phi(\mathbf{x})}$. This objective encourages the policy to sample diverse, high-reward trajectories in proportion to their rewards, rather than collapsing to dominant modes as in standard reward maximization.

While the KL-based formulation provides a principled target distribution, we derive a more practical, RL-style objective that facilitates efficient policy optimization.
\begin{proposition}\label{prop:kl_gflownet_equivalence}
In terms of expected gradients, minimizing the KL objective in Eq.~\ref{eq:kl_objective} is equivalent to minimizing the trajectory balance loss used in GFlowNet~\citep{malkin2022trajectory,malkin2022gfnhvi,lee2024learning,bartoldson2025trajectory}:
\begin{align}
\min_\theta \mathcal{D}_{\mathrm{KL}} \left( 
\pi_\theta(\mathbf{y} \mid \mathbf{x}) \,\middle\|\, 
\frac{\exp(\beta r(\mathbf{x}, \mathbf{y}))}{Z_\phi(\mathbf{x})}
\right)
\quad \Longleftrightarrow \quad 
\min_\theta 
\underbrace{
\left( \log Z_\phi(\mathbf{x}) + \log \pi_\theta(\mathbf{y} \mid \mathbf{x}) - \beta r(\mathbf{x}, \mathbf{y}) \right)^2
}_{\text{Trajectory Balance}}\label{eq:kl_tb}
\end{align}
\end{proposition}
\begin{remark}[\textit{Trajectory balance as a practical surrogate for KL minimization}]
Given the equivalence established in Proposition~\ref{prop:kl_gflownet_equivalence}, the KL-based distribution matching objective can be reformulated as the trajectory balance loss. This reformulation provides a practical optimization approach by using a stable squared loss form rather than direct KL optimization, and by treating $Z_\phi(\mathbf{x})$ as a learnable parameter rather than requiring explicit computation of the intractable partition function. The trajectory balance objective thus serves as a tractable surrogate for reward-guided KL minimization that can be directly integrated into existing RL frameworks.
\end{remark}

% Having established a KL-based objective that aligns the policy with the full reward distribution in theory, we turn to the practical challenges that arise during its application. 

\subsection{FlowRL}\label{sec:flowrl}
As established in Proposition~\ref{prop:kl_gflownet_equivalence}, the target reward distribution can be approximated by optimizing the trajectory balance objective. However, applying this objective directly to long CoT reasoning introduces two key challenges:

\paragraph{Problem I: Exploding gradients from long trajectories.} Trajectory balance is a sequence-level objective, and applying it to long CoT reasoning with up to 8K tokens leads to exploding gradients and unstable updates. This issue is not observed in prior GFlowNets works, which typically operate on short trajectories in small discrete spaces. Specifically, the log-probability term $\log \pi_\theta(\mathbf{y} \mid \mathbf{x})$ decomposes into a token-wise sum, $\sum_{t} \log \pi_\theta(\mathbf{y}_t \mid \mathbf{y}_{<t}, \mathbf{x})$, causing the gradient norm to potentially scale with sequence length.

\paragraph{Problem II: Sampling mismatch.} Mainstream RL algorithms such as PPO and GRPO commonly perform micro-batch updates and reuse trajectories collected from an old policy $\pi_{\theta_{\text{old}}}$, enabling data-efficient training. In contrast, the KL-based trajectory balance objective assumes fully on-policy sampling, where responses are drawn from the current policy. This mismatch poses practical limitations when integrating trajectory balance into existing RL pipelines.

These limitations motivate our reformulation that retains the benefits of distribution matching while addressing key practical challenges. To enable this reformulation, we first redefine the reward function following established practices in GFlowNets literature~\citep{lee2024learning,bartoldson2025trajectory,yuflow} by incorporating a reference model as a prior constraint on the reward distribution. Specifically, we modify the original $\exp(\beta r(\mathbf{x}, \mathbf{y}))$ to include the reference model: 
\begin{equation}
\exp\left(\beta\; r(\mathbf{x}, \mathbf{y})\right) \cdot \pi_{\mathrm{ref}}(\mathbf{y} \mid \mathbf{x}),
\label{eq:redefined_reward}
\end{equation}
where $r(\mathbf{x}, \mathbf{y})$ denotes the outcome reward commonly used in reinforcement learning and $\pi_{\mathrm{ref}}$ is the initial pre-trained model. We follow~\cite{guo2025deepseek} to use outcome-based reward signals, and apply group normalization to $r(\mathbf{x}, \mathbf{y})$ as $\hat{r}_i = (r_i - \mathrm{mean}(\mathbf{r}))/\mathrm{std}(\mathbf{r})$, where $\mathbf{r} = \{r_1, r_2, \dots, r_G\}$ denotes the set of rewards within a sampled group. By substituting the redefined reward formulation Eq.~\ref{eq:redefined_reward} into Eq.~\ref{eq:kl_tb}, we derive the following objective\footnote{The substitution replaces $\beta r(\mathbf{x}, \mathbf{y})$
in trajectory balance objective Eq.~\ref{eq:kl_tb} with $\beta r(\mathbf{x}, \mathbf{y}) + \log \pi_{\mathrm{ref}}(\mathbf{y} \mid \mathbf{x})$ to incorporate the reference model constraint.}:
\begin{align}
\min_\theta \left(\log Z_\phi(\mathbf{x}) + \log \pi_\theta(\mathbf{y} \mid \mathbf{x})-\beta \; \hat{r}_i(\mathbf{x}, \mathbf{y}) -\log\pi_{\mathrm{ref}}(\mathbf{y} \mid \mathbf{x}) \right)^2 \label{eq:gfn}
\end{align}

\begin{remark}[\textit{Reward shaping via length normalization}]
Trajectory balance treats both the initial flow and the outcome reward as sequence-level quantities. In contrast, standard policy optimization methods such as PPO or GRPO assign rewards at the token level and compute gradients at each step. However, for trajectories of varying lengths (\eg, CoT responses), this mismatch can cause the log-probability term $\log \pi_\theta(\mathbf{y} \mid \mathbf{x}) = \sum_{t=1}^{|\mathbf{y}|} \log \pi_\theta(y_t \mid y_{<t}, \mathbf{x})$ to scale with sequence length. To address this, we apply a form of reward shaping by normalizing log-probabilities with respect to sequence length. Specifically, we rescale the term as $\frac{1}{|\mathbf{y}|} \log \pi_\theta(\mathbf{y} \mid \mathbf{x})$, balancing the contributions of long and short sequences and stabilizing the learning signal.
\end{remark}

\begin{remark}[\textit{Importance sampling for data-efficient training}]
To mitigate sampling mismatch, we employ importance sampling inspired by PPO to stabilize policy updates with off-policy data. We re-weight stale trajectories using the importance ratio $w = \pi_\theta(\mathbf{y}\mid\mathbf{x})/\pi_{\text{old}}(\mathbf{y}\mid\mathbf{x})$, which serves as a coefficient in the surrogate loss. Since our objective focuses on optimizing trajectory balance rather than expected return, we detach the gradient from the current policy to prevent excessive policy drift: $w = \text{detach}[\pi_\theta(\mathbf{y} \mid \mathbf{x})] / \pi_{\text{old}}(\mathbf{y} \mid \mathbf{x})$. For additional stability, we incorporate PPO-style clipping to bound the importance weights: $w = \text{clip}\left(\frac{\pi_\theta(\mathbf{y} \mid \mathbf{x})}{\pi_{\text{old}}(\mathbf{y} \mid \mathbf{x})}, 1 - \epsilon, 1 + \epsilon\right)^{\text{detach}}.$
\end{remark}

Incorporating these improvements into Eq.~\ref{eq:gfn}, we arrive at the following FlowRL objective:

\begin{tcolorbox}[
  enhanced,
  breakable,
  float,
  floatplacement=h!,
  title=\textbf{FlowRL},
  colframe=myblue,
  colback=myblue!8,
  coltitle=white,
  parbox=false,
  left=8pt,
  right=8pt,
  top=6pt,
  bottom=6pt,
  grow to left by=3pt,
  grow to right by=3pt,
  toprule=2pt,
  titlerule=1pt,
  leftrule=1pt,
  rightrule=1pt,
  bottomrule=1pt,
]
\begin{equation}
% \resizebox{\linewidth}{!}{$
\mathcal{L}_{\text{FlowRL}} =
w \cdot \left( 
\log Z_\phi(\mathbf{x}) 
+ \frac{1}{|\mathbf{y}|} \log \pi_\theta(\mathbf{y} \mid \mathbf{x}) 
- \beta \hat{r}(\mathbf{x}, \mathbf{y}) 
- \frac{1}{|\mathbf{y}|}\log \pi_{\mathrm{ref}}(\mathbf{y} \mid \mathbf{x})
\right)^2 
% $}
\label{eq:flowrl}
\end{equation}
\end{tcolorbox}

where the clipped importance weight $w$ and normalized reward $\hat{r}(\mathbf{x}, \mathbf{y})$ are defined as:
\begin{align}
w = \text{clip}(\frac{\pi_\theta(\mathbf{y} \mid \mathbf{x})}{\pi_{\text{old}}(\mathbf{y} \mid \mathbf{x})}
, 1 - \epsilon, 1 + \epsilon)^{\text{detach}},
\quad
\hat{r}_i &= \frac{r_i - \mathrm{mean}(\mathbf{r})}{\mathrm{std}(\mathbf{r})}.
\end{align}
We use this objective to update the policy parameters $\theta$ during training, and refer to this strategy as \textit{FlowRL}. Implementation details and theoretical analysis are provided in~\S~\ref{sec:setup_models} and~\S~\ref{sec:theoretical_analysis}, respectively.

\section{Experiment Settings}\label{sec:experiments}
\paragraph{Backbone Models.}\label{sec:setup_models} 
There are two learnable modules in Eq.~\ref{eq:flowrl}: the policy model $\pi_\theta$ and the partition function $Z_\phi$. For the policy model $\pi_\theta$, we use \texttt{Qwen-2.5-7B/32B}~\citep{qwen2.5} for math tasks and \texttt{DeepSeek-R1-Distill-Qwen-7B}~\citep{deepseekai2025deepseekr1incentivizingreasoningcapability} for code tasks, respectively. The reference model $\pi_{\text{ref}}$ is the corresponding fixed pretrained model. 
For partition function $Z_\phi$, following~\cite{lee2024learning}, we use a randomly initialized 3-layer MLP with hidden dimensions matching those of the base model. The input to $Z_\phi$ is the mean of the language model's hidden states after encoding the input $\mathbf{x}$, and the output is a scalar value. We detail the implementation of $Z_\phi$ in \S~\ref{sec:partition_function}.
All training scripts are based on the {veRL}~\citep{verl}. For the reward function, following~\cite{lee2024learning}, we set the hyperparameter $\beta = 15$. 

\paragraph{Baselines.} 
We compare our method against three representative reward-maximization RL baselines: REINFORCE++~(R++; \citealp{NIPS1999_464d828b,hu2501reinforce++}), PPO~\citep{schulman2017proximal}, and GRPO~\citep{shao2024deepseekmath}. All baselines follow the official veRL recipes, with consistent training configurations. For fair comparison, all methods use the same learning rate, batch size, and training steps, and are evaluated at convergence using identical step counts.

\paragraph{Training Configuration.} We experiment on both math and code domains. For the math domain, we use the training set collected from DAPO~\citep{yu2025dapo}. For the code domain, we follow the setup of DeepCoder~\citep{deepcoder2025}, using their training set. For 7B model training, we use a single node equipped with 8 NVIDIA H800 GPUs (80GB memory each). For 32B model training, we scale to 4 nodes with 32 GPUs to accommodate the larger memory requirements. All experiments use \texttt{max\_prompt\_length} = 2048 and \texttt{max\_response\_length} = 8192 across both model sizes. We use a batch size of 512 for math reasoning tasks and 64 for code reasoning tasks. We set the learning rate to 1e-6 and enable dynamic batch sizing in veRL for efficient training. For GRPO and FlowRL, we configure \texttt{rollout\_n} = 8, meaning each prompt generates 8 response rollouts as the group size.

\paragraph{Evaluation Configuration.} For the math domain, we evaluate on six challenging benchmarks: AIME 2024/2025~\citep{AIME}, AMC 2023~\citep{AMC}, MATH-500~\citep{lightman2023lets}, Minerva~\citep{minerva}, and Olympiad~\citep{he2024olympiadbench}. For the code domain, we evaluate on LiveCodeBench~\citep{jain2024livecodebench}, CodeForces~\citep{penedo2025codeforces}, and HumanEval+~\citep{chen2021evaluating}. For all evaluation datasets, we perform 16 rollouts and report the average accuracy, denoted as Avg@16. We further report rating and percentile for Codeforces. During generation, we use sampling parameters of \texttt{temperature} = 0.6 and \texttt{top\_p} = 0.95 for all evaluations. The response length for evaluation is set to 8,192, consistent with the training configuration.

\begin{table*}[t!]
  \centering
% \vspace{5pt}
  \resizebox{\textwidth}{!}{ 
  \begin{tabular}{l|llllll|l}
      \toprule[1.25pt]
        \textbf{Models} 
        & \textbf{AIME24} 
        & \textbf{AIME25}
        & \textbf{AMC23}
        & \textbf{MATH500}
        & \textbf{Minerva}
        & \textbf{Olympiad}
        & \textbf{Avg} \\
      \midrule[1.1pt]
      \rowcolor[rgb]{0.93,0.93,0.93}
      \multicolumn{8}{c}{\texttt{Qwen2.5-32B-Base, Max Response Len = 8K tokens}} \\
      \textcolor{gray}{\texttt{Backbone}} & \textcolor{gray}{4.58} & \textcolor{gray}{2.08} & \textcolor{gray}{28.59} & \textcolor{gray}{52.48} & \textcolor{gray}{26.99} & \textcolor{gray}{21.37} & \textcolor{gray}{22.68} \\
      R++ & $14.79_{\textcolor{scoregreen}{+10.21}}$ & $9.17_{\textcolor{scoregreen}{+7.08}}$ & $52.65_{\textcolor{scoregreen}{+24.06}}$ & $44.35_{\textcolor{red}{-8.13}}$ & $17.37_{\textcolor{red}{-9.62}}$ & $24.52_{\textcolor{scoregreen}{+3.15}}$ & 27.14 \\
      PPO & $26.87_{\textcolor{scoregreen}{+22.29}}$ & $20.41_{\textcolor{scoregreen}{+18.33}}$ & $76.40_{\textcolor{scoregreen}{+47.81}}$ & $69.17_{\textcolor{scoregreen}{+16.69}}$ & $28.79_{\textcolor{scoregreen}{+1.80}}$ & $37.90_{\textcolor{scoregreen}{+16.53}}$ & 43.25 \\
      GRPO & $23.12_{\textcolor{scoregreen}{+18.54}}$ & $14.58_{\textcolor{scoregreen}{+12.50}}$ & $76.87_{\textcolor{scoregreen}{+48.28}}$ & $61.60_{\textcolor{scoregreen}{+9.12}}$ & $18.95_{\textcolor{red}{-8.04}}$ & $34.94_{\textcolor{scoregreen}{+13.57}}$ & 38.34 \\
      \midrule
      FlowRL & $23.95_{\textcolor{scoregreen}{+19.37}}$ & $21.87_{\textcolor{scoregreen}{+19.79}}$ & $73.75_{\textcolor{scoregreen}{+45.16}}$ & $80.75_{\textcolor{scoregreen}{+28.27}}$ & $38.21_{\textcolor{scoregreen}{+11.22}}$ & $51.83_{\textcolor{scoregreen}{+30.46}}$ & \textbf{48.39} \\
      \midrule
      \rowcolor[rgb]{0.93,0.93,0.93}
      \multicolumn{8}{c}{\texttt{Qwen2.5-7B-Base, Max Response Len = 8K tokens}} \\
      \textcolor{gray}{\texttt{Backbone}} & \textcolor{gray}{4.38} & \textcolor{gray}{2.08} & \textcolor{gray}{30.78} & \textcolor{gray}{54.47} & \textcolor{gray}{22.38} & \textcolor{gray}{24.03} & \textcolor{gray}{23.02} \\
      R++ & $11.04_{\textcolor{scoregreen}{+6.66}}$ & $5.41_{\textcolor{scoregreen}{+3.33}}$ & $66.71_{\textcolor{scoregreen}{+35.93}}$ & $54.25_{\textcolor{red}{-0.22}}$ & $24.37_{\textcolor{scoregreen}{+1.99}}$ & $27.33_{\textcolor{scoregreen}{+3.30}}$ & 31.52 \\
      PPO & $9.38_{\textcolor{scoregreen}{+5.00}}$ & $7.29_{\textcolor{scoregreen}{+5.21}}$ & $63.43_{\textcolor{scoregreen}{+32.65}}$ & $57.98_{\textcolor{scoregreen}{+3.51}}$ & $26.53_{\textcolor{scoregreen}{+4.15}}$ & $27.25_{\textcolor{scoregreen}{+3.22}}$ & 31.98 \\
      GRPO & $13.54_{\textcolor{scoregreen}{+9.16}}$ & $9.79_{\textcolor{scoregreen}{+7.71}}$ & $64.53_{\textcolor{scoregreen}{+33.75}}$ & $57.05_{\textcolor{scoregreen}{+2.58}}$ & $23.06_{\textcolor{scoregreen}{+0.68}}$ & $26.88_{\textcolor{scoregreen}{+2.85}}$ & 32.48 \\
      \midrule
      FlowRL & $15.41_{\textcolor{scoregreen}{+11.03}}$ & $10.83_{\textcolor{scoregreen}{+8.75}}$ & $54.53_{\textcolor{scoregreen}{+23.75}}$ & $66.96_{\textcolor{scoregreen}{+12.49}}$ & $31.41_{\textcolor{scoregreen}{+9.03}}$ & $34.61_{\textcolor{scoregreen}{+10.58}}$ & \textbf{35.63} \\
      \bottomrule[1.25pt]
  \end{tabular}}
    \caption{\textbf{Results on math reasoning benchmarks.} We report Avg@16 accuracy with relative improvements shown as subscripts. Positive gains are shown in \textcolor{scoregreen}{green} and negative changes in \textcolor{red}{red}. FlowRL outperforms all baselines across both 7B and 32B model scales.}
\label{tab:math_results}
% \vspace{-10pt}
\end{table*}

\begin{table*}[t!]
  \centering
  \resizebox{0.85\textwidth}{!}{  % Reduced from \textwidth to 0.85\textwidth
  \begin{tabular}{l|ll|ll|l}
    \toprule[1.25pt]
    \textbf{Models} 
    & \multicolumn{2}{c|}{\textbf{LiveCodeBench}} 
    & \multicolumn{2}{c|}{\textbf{CodeForces}} 
    & \multicolumn{1}{c}{\textbf{HumanEval+}} \\  % Added \multicolumn{1}{c} to ensure proper alignment
    \cmidrule(lr){2-3} \cmidrule(lr){4-5} \cmidrule(lr){6-6}  % Added cmidrule for HumanEval
    & Avg@16 & Pass@16
    & Rating & Percentile
    & Avg@16 \\
    \midrule[1.1pt]
    \rowcolor[rgb]{0.93,0.93,0.93}
      \multicolumn{6}{c}{\texttt{DeepSeek-R1-Distill-Qwen-7B, Max Response Len = 8K tokens}}  \\
    \textcolor{gray}{Backbone}
    & \textcolor{gray}{30.68} & \textcolor{gray}{49.46}
    & \textcolor{gray}{886.68} & \textcolor{gray}{19.4\%}
    & \textcolor{gray}{80.90} \\
    
    R++ 
    & $30.46_{\textcolor{red}{-0.22}}$ & $52.68_{\textcolor{scoregreen}{+3.22}}$ 
    & $1208.03_{\textcolor{scoregreen}{+321.35}}$ & $56.8\%_{\textcolor{scoregreen}{+37.4\%}}$ 
    & $76.61_{\textcolor{red}{-4.29}}$ \\
    
    PPO               
    & $35.10_{\textcolor{scoregreen}{+4.42}}$ & $54.48_{\textcolor{scoregreen}{+5.02}}$
    & $1403.07_{\textcolor{scoregreen}{+516.39}}$ & $73.7\%_{\textcolor{scoregreen}{+54.3\%}}$
    & $82.32_{\textcolor{scoregreen}{+1.42}}$ \\
    
    GRPO              
    & $32.75_{\textcolor{scoregreen}{+2.07}}$ & $52.32_{\textcolor{scoregreen}{+2.86}}$
    & $1313.82_{\textcolor{scoregreen}{+427.14}}$ & $67.1\%_{\textcolor{scoregreen}{+47.7\%}}$
    & $80.13_{\textcolor{red}{-0.77}}$ \\
    \midrule
    FlowRL            
    & $\mathbf{37.43}_{\textcolor{scoregreen}{+6.75}}$ & $\mathbf{56.27}_{\textcolor{scoregreen}{+6.81}}$
    & $\mathbf{1549.47}_{\textcolor{scoregreen}{+662.79}}$ & $\mathbf{83.3\%}_{\textcolor{scoregreen}{+63.9\%}}$
    & $\mathbf{83.28}_{\textcolor{scoregreen}{+2.38}}$ \\
    \bottomrule[1.25pt]
  \end{tabular}}
    \caption{\textbf{Results on code benchmarks.} We report metrics with relative improvements shown as subscripts. Positive gains are shown in \textcolor{scoregreen}{green} and negative changes in \textcolor{red}{red}. FlowRL achieves the strongest performance across all three benchmarks.} 
% \vspace{5pt}
\label{tab:code_results}
% \vspace{-10pt}
\end{table*}

\section{Results}\label{sec:results}
\subsection{Main Results}
Our experimental results, summarized in Table~\ref{tab:math_results} and Table~\ref{tab:code_results}, demonstrate that FlowRL consistently outperforms all reward-maximization baselines across both math and code reasoning domains.  Table~\ref{tab:math_results} reports results on math reasoning benchmarks using both 7B and 32B base models, while Table~\ref{tab:code_results} presents the corresponding results on code reasoning tasks. On math reasoning tasks, FlowRL achieves the highest average accuracy of 35.6\% with the 7B model and 48.4\% with the 32B model, surpassing PPO by 5.1\% and GRPO by 10.1\% on the 32B model. FlowRL shows strong improvements on challenging benchmarks like MATH-500 and Olympiad problems, demonstrating consistent gains across diverse mathematical domains. On code generation tasks, FlowRL achieves compelling improvements with the highest Avg@16 score of 37.43\% on LiveCodeBench, a Codeforces rating of 1549.47 with 83.3\% percentile ranking, and 83.28\% accuracy on HumanEval+, outperforming all baselines across the board. 
These consistent performance gains across both domains and model scales provide strong empirical evidence that FlowRL's flow-balanced optimization successfully enhances generalization. This improvement comes from promoting diverse solution exploration compared to previous reward-maximizing RL approaches.

\begin{table*}[t]
\centering
\resizebox{\textwidth}{!}{ 
\begin{tabular}{l|cccccc|c}
    \toprule[1.25pt]
      \textbf{Method} 
      & \textbf{AIME 2024} 
      & \textbf{AIME 2025}
      & \textbf{AMC 2023}
      & \textbf{MATH-500}
      & \textbf{Minerva}
      & \textbf{Olympiad}
      & \textbf{Avg} \\
    \midrule[1.1pt]
    FlowRL & 15.41 & 10.83 & 54.53 & 66.96 & 31.41 & 34.61 & 35.63 \\
    \quad w/o IS & 6.25 & 7.91 & 41.40 & 56.97 & 22.19 & 25.52 & 26.71 \\
    \midrule
    \cite{zhang2025improving} & 10.41 & 6.66 & 53.75 & 66.50 & 30.97 & 33.72 & 33.67 \\
    \bottomrule[1.25pt]
\end{tabular}}
\caption{Ablation study on FlowRL with Qwen2.5-7B as the base model. Avg@16 accuracy is reported across six math reasoning benchmarks. IS denotes importance sampling.}
\label{tab:ablation_flowrl}
\end{table*}

\subsection{Ablation Studies} 
\begin{wrapfigure}{r}{0.45\textwidth}
    % \vspace{-2em}
    \centering
    \begin{center}
    \includegraphics[width=\linewidth]{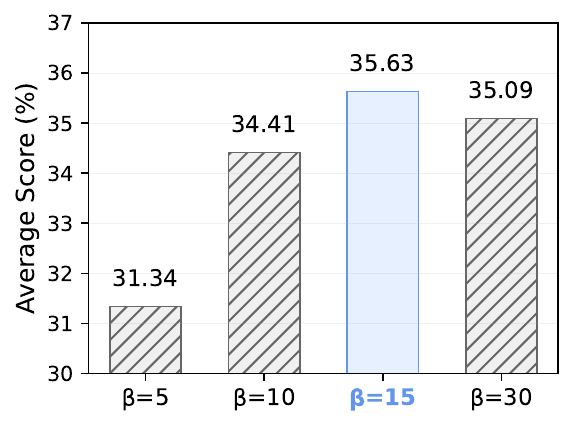} 
        \vspace{-2em}
        \caption{Ablation study on the $\beta$ in FlowRL. $\beta = 15$ (highlighted in blue) achieves the best performance.}\label{fig:ablation_on_beta}
    \end{center}
    \vspace{-1em}
\end{wrapfigure}
We conduct ablation studies on importance sampling and the $\beta$ hyperparameter. For importance sampling, we compared the performance with and without it, and implemented a combined loss approach proposed by \cite{zhang2025improving} that simultaneously optimizes both GFlowNets and PPO objectives. This combined loss focuses on optimizing diffusion models, and we adapt it to long CoT reasoning tasks for comparison. Table~\ref{tab:ablation_flowrl} demonstrates that importance sampling substantially improves FlowRL performance across all math reasoning benchmarks. Compared to~\cite{zhang2025improving}, using importance sampling as a trajectory-level ratio is more suitable than the combined loss of GFlowNets and PPO. 
The performance drop without importance sampling (from 35.63\% to 26.71\%) highlights the critical role of correcting for distribution mismatch between rollout generation and policy training. 
For the hyperparameter $\beta$, we conduct a series of parameter ablation studies, and Figure~\ref{fig:ablation_on_beta} shows that $\beta = 15$ achieves optimal performance, with detailed results shown in Table~\ref{tab:ablation_flowl}.

\section{Analysis}
\subsection{Diversity Analysis}
To assess solution diversity, we follow the approach of~\cite{yuflow} and employ \texttt{GPT-4o-mini}~\citep{openai2024gpt4omini} to evaluate all responses generated by each method on AIME 24/25. The evaluation prompt is shown in Appendix~\ref{box:diversity-eval}. As shown in Figure~\ref{fig:diversity_scores}, FlowRL achieves higher diversity scores compared to baseline methods. This demonstrates that FlowRL improves sample diversity compared to baselines, which tend to exhibit repetitive solution patterns. 
This diversity evaluation reveals significant differences in exploration patterns across methods. This nearly doubling of diversity score compared to the strongest baseline (PPO) indicates that FlowRL generates qualitatively different solution approaches rather than minor variations of the same strategy. The diversity analysis provides empirical validation of our core hypothesis that flow-balanced optimization promotes mode coverage in complex reasoning tasks. 

\subsection{Case Study}

\begin{wrapfigure}{r}{0.45\textwidth}
    \vspace{-2em}
    \centering
    \begin{center}
        \includegraphics[width=\linewidth]{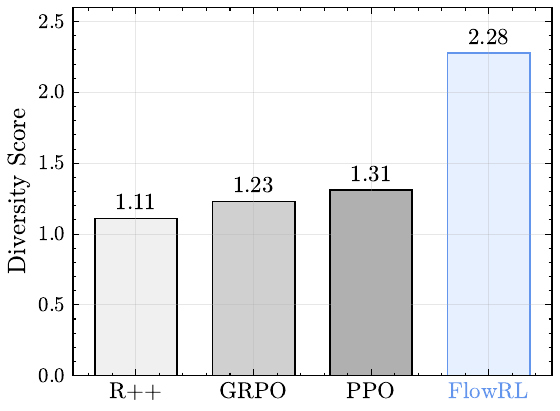} 
        \vspace{-2em}
        \caption{GPT-judged diversity scores on rollouts of AIME 24/25 problems. FlowRL generates more diverse solutions than R++, GRPO, and PPO.}\label{fig:diversity_scores}
    \end{center}
    \vspace{-2em}
\end{wrapfigure}
Table~\ref{tab:case_study} illustrates the behavioral differences between GRPO and FlowRL on a representative AIME problem. GRPO exhibits repetitive patterns, applying AM-GM three times and getting stuck in identity loops, failing to solve the problem. FlowRL explores more diverse actions: it sets $a=b$, derives a cubic equation, finds the rational root, and reaches the correct answer. This shows that FlowRL successfully avoids the repetitive exploration patterns. The contrast reveals fundamental differences in exploration strategies: GRPO's reward-maximizing approach leads to exploitation of familiar techniques (AM-GM inequality) without exploring alternatives, eventually reaching contradictory conclusions like $a=b=c$. In contrast, FlowRL's distribution-matching enables strategic decisions such as the symmetry assumption $a=b$, which transforms the problem into a tractable cubic equation $a^3-27a+46=0$, allowing systematic solution through rational root testing and polynomial factorization.

\begin{table*}[t]
\centering
\caption{Case study comparing GRPO and FlowRL rollouts on an AIME problem. GRPO exhibits repetitive patterns (AM-GM \(\times 3\), identity loops \(\times 2\)), while FlowRL follows a more diverse solution path.}
\resizebox{0.98\textwidth}{!}{ 
\begin{tabular}{@{}p{0.14\linewidth}p{0.82\linewidth}@{}}
\toprule
 & \textbf{Content (boxed = actions; “\(\times k\)” = repeated; “…” = omitted)} \\
\midrule
\textbf{Question} &
Let $\mathcal{B}$ be the set of rectangular boxes with surface area $54$ and volume $23$.
Let $r$ be the radius of the smallest sphere that can contain each box in $\mathcal{B}$.
If $r^2=\frac{p}{q}$ with $\gcd(p,q)=1$, find $p+q$. \\[4pt]
\midrule
\rowcolor{GRPOrow}
\textbf{\textcolor{GRPOGray}{GRPO}} &
“… denote $a,b,c$ … 
\gact{$2(ab{+}bc{+}ca)=54$, $abc=23$}
… \gact{$d=\sqrt{a^2+b^2+c^2}$, $r=d/2$}
… \gact{$(a{+}b{+}c)^2 = a^2{+}b^2{+}c^2 + 2(ab{+}bc{+}ca)$}
… \gact{AM--GM}\repbadgeR{3}:
\gmini{AM--GM (1)} … \gmini{AM--GM (2)} … \gmini{AM--GM (3)}
… \gact{$(a{+}b{+}c)^3$ identity loop}\repbadgeR{2}:
\gmini{loop (1)} … \gmini{loop (2)}
… \gact{$a=b=c$ (contradiction)}
… \gact{back to $(a{+}b{+}c)^2$}
… no factorization …” \\[6pt]
\midrule
\rowcolor{FlowRLrow}
\textbf{\textcolor{FlowBlue}{FlowRL}} &
“… let $a,b,c$ with \fact{$2(ab{+}bc{+}ca)=54$, $abc=23$}
… \fact{$d=\sqrt{a^2+b^2+c^2}$, $r=d/2$}
… \fact{$(a{+}b{+}c)^2 \Rightarrow a^2{+}b^2{+}c^2=s^2-54$}
… \fact{$a=b$}
… \fact{$a^3-27a+46=0$}
… \fact{rational root $a=2$}
… \fact{factor $(a-2)(a^2+2a-23)$}
… \fact{branch $a=-1+2\sqrt6$}
… \fact{back-sub $c=23/a^2$}
… \fact{$a^2{+}b^2{+}c^2=\frac{657}{16}$}
… \fact{$r^2=\frac{657}{64}$}
… \fact{Answer 721} …” \\
\bottomrule
\end{tabular}}
\label{tab:case_study}
\end{table*}

\section{Related Work}\label{sec:related_work}
Our work relates to GFlowNets, Flow-Matching Policies, Length Normalization and KL Regularization. We discuss three topics that relate most closely to our work in this section, and the other topics are included in Appendix~\ref{sec:more_related_work}. 

\paragraph{Reinforcement Learning for LLM Reasoning.}
RL has emerged as a powerful approach for LLM post-training on reasoning tasks~\citep{NIPS1999_464d828b,schulman2017proximal,lightman2023let,shao2024deepseekmath,guo2025deepseek}. 
Most approaches employ
reward-maximizing RL to optimize expected cumulative returns. Entropy regularization~\citep{haarnoja2018soft,ahmed2019understanding,cheng2025reasoning} is a classical technique for mitigating mode collapse by promoting diversity in the policy’s output distribution, and has also been shown to enhance reasoning capabilities in various settings~\citep{eysenbach2021maximum,chao2024maximum}.
However, for long CoT reasoning, the extended trajectory length (e.g., more than 8k tokens) makes it difficult for the regularization signal to effectively influence reward-maximizing learning. Recent work~\citep{cheng2025reasoning,wang2025beyond,cui2025entropy,dong2025agentic} has discovered that training with more diverse or high-entropy training data can further enhance training effectiveness. Compared to traditional entropy regularization, the above methods explicitly increase the proportion of low-probability (i.e., high-entropy) tokens in the training data. In our work, we address the mode-collapse problem by fundamentally shifting from reward maximization to reward distribution matching in our RL formulation. 
See Appendix~\ref{sec:more_related_work} for detailed comparisons.

\paragraph{GFlowNets.}
GFlowNets~\citep{JMLR:v24:22-0364} represent a class of diversity-driven algorithms designed to balance probability flows across states. 
They have rich connections to probabilistic modeling methods~\citep{zhang2022unifying,zhang2022generative,zhang2024diffusion,Zimmermann2022AVP,malkin2022gfnhvi,MaBakingSI}, and control methods~\citep{pan2023better,pan2022gafn,pan2023stochastic,zhang2024distributional,tiapkin2024generative}.
This advantage has enabled GFlowNets to achieve successful applications in multiple downstream tasks, such as molecular drug discovery~\citep{jain2022biological,jain2022multiobjective,Liu2022GFlowOutDW,Jain2023GFlowNetsFA,Shen2023TowardsUA,pan2023pretraining,Kim2023LocalSG,kim2024learning}, phylogenetic inference~\citep{zhou2024phylogfn}, and combinatorial optimization~\citep{Zhang2023RobustSW,Zhang2023LetTF}.
For generative AI,
GFlowNets provide a powerful approach to align pretrained models in scenarios such as image generation~\citep{zhang2025improving,yun2025learning} and language model fine-tuning~\citep{hu2024amortizing,yuflow,lee2024learning}.
Another line of work primarily focuses on the theoretical aspects of GFlowNets. Recent theoretical studies have interpreted GFlowNets as solving a maximum entropy reinforcement learning problem within a modified Markov Decision Process (MDP)~\citep{tiapkin2024generative,deleu2024discrete,mohammadpour2024maximum}. 
These theoretical contributions have inspired us to enhance reinforcement learning from a more foundational standpoint using GFlowNets principles. A comprehensive overview of GFlowNets theory can be found in Appendix~\ref{sec:gflownets_notation}.

\paragraph{Flow-Matching Policies.}
Flow matching simplifies diffusion-based approaches by learning vector fields that transport samples from prior to target distributions~\citep{lipman2023flow}. Recent work has explored flow matching for policy optimization. \cite{mcallister2025flow} reformulates policy optimization using advantage-weighted ratios from conditional flow matching loss, enabling flow-based policy training without expensive likelihood computations. \cite{pfrommer2025reinforcement} explored reward-weighted flow matching for improving policies beyond demonstration performance. \cite{park2025flow} uses a separate one-step policy to avoid unstable backpropagation through time when training flow policies with RL. \cite{zhang2025improving} proposed a combined loss function integrating PPO and GFlowNets to optimize diffusion model alignment.~\cite{lv2025flow} integrates flow-based policy representation with Wasserstein regularized optimization for online reinforcement learning.
However, these approaches focus on continuous control, image generation, or vision-action models, rather than addressing mode-collapse limitations in reward-maximizing RL. Inspired by flow matching principles, our work improves upon RL training to enhance training stability while promoting diverse solution exploration.

\section{Conclusion}
In this work, we introduce FlowRL, which transforms scalar rewards into normalized target distributions using a learnable partition function and minimizes the reverse KL divergence between the policy and target distribution. We demonstrate that this approach is theoretically equivalent to trajectory balance objectives from GFlowNets and implicitly maximizes both reward and entropy, thereby promoting diverse reasoning trajectories. To further address gradient explosion and sampling mismatch issues in long CoT reasoning, we incorporate importance sampling and length normalization. Through experiments on math and code reasoning benchmarks, FlowRL achieves consistent improvements across all tasks compared to GRPO and PPO. Our diversity analysis and case studies confirm that FlowRL generates more varied solution approaches while avoiding repetitive patterns.

\section*{Acknowledgments}

We are grateful to Mingqian Feng and Yuetai Li for their valuable discussions and feedback, which helped improve the quality of this work. 

% \bibliography{iclr2026_conference}
% \bibliographystyle{iclr2026_conference}
% \newpage
% \addcontentsline{toc}{section}{References}
\bibliography{flowrl}

\newpage
\appendix

\section{Proof of Proposition~\ref{prop:kl_gflownet_equivalence}}\label{sec:proposition_1_proof}
We begin by analyzing the gradient of the Kullback–Leibler (KL) divergence between the policy $\pi_\theta(\mathbf{y} \mid \mathbf{x})$ and the target reward distribution $\frac{\exp(\beta r(\mathbf{x},\mathbf{y}))}{Z_\phi(\mathbf{x})}$:
\begin{equation}
\resizebox{0.9\linewidth}{!}{$
\begin{split}
&\nabla_\theta D_{\mathrm{KL}} \left(
\pi_\theta(\mathbf{y} \mid \mathbf{x}) \;\Vert\; 
\frac{\exp(\beta r(\mathbf{x},\mathbf{y}))}{Z_\phi(\mathbf{x})}\right) \\
&= \nabla_\theta \int \pi_\theta(\mathbf{y} \mid \mathbf{x}) \log 
\left[
\frac{\pi_\theta(\mathbf{y} \mid \mathbf{x}) \cdot Z_\phi(\mathbf{x})}{\exp(\beta r(\mathbf{x},\mathbf{y}))}\right] d\mathbf{y} \\
&= \int \nabla_\theta \pi_\theta(\mathbf{y} \mid \mathbf{x}) \log 
\left[\frac{Z_\phi(\mathbf{x})\pi_\theta(\mathbf{y} \mid \mathbf{x}) }{\exp(\beta r(\mathbf{x},\mathbf{y}))}\right] d\mathbf{y} + \int \pi_\theta(\mathbf{y} \mid \mathbf{x}) \nabla_\theta \log 
\left[
\frac{Z_\phi(\mathbf{x})\pi_\theta(\mathbf{y} \mid \mathbf{x}) 
}{\exp(\beta r(\mathbf{x},\mathbf{y}))}\right] d\mathbf{y} \\
&= \int \pi_\theta(\mathbf{y} \mid \mathbf{x})\, \nabla_\theta \log\pi_\theta(\mathbf{y} \mid \mathbf{x})\, \log 
\left[
\frac{Z_\phi(\mathbf{x})\pi_\theta(\mathbf{y} \mid \mathbf{x}) 
}{\exp(\beta r(\mathbf{x},\mathbf{y}))}\right] \, d\mathbf{y} 
+ \underbrace{
\int \pi_\theta(\mathbf{y} \mid \mathbf{x})\, \nabla_\theta \log\pi_\theta(\mathbf{y} \mid \mathbf{x})\, d\mathbf{y}
}_{= \nabla_\theta \int \pi_\theta(\mathbf{y} \mid \mathbf{x})\, d\mathbf{y} = \nabla_\theta 1 = 0} \\
&= \int \pi_\theta(\mathbf{y} \mid \mathbf{x})\, \nabla_\theta \log\pi_\theta(\mathbf{y} \mid \mathbf{x})\, \log 
\left[\frac{Z_\phi(\mathbf{x})\pi_\theta(\mathbf{y} \mid \mathbf{x}) }{\exp(\beta r(\mathbf{x},\mathbf{y}))}\right] \, d\mathbf{y} \\
&= \mathbb{E}_{\mathbf{y} \sim \pi_\theta(\cdot \mid \mathbf{x})} 
\left[ 
\log 
\left(
\frac{Z_\phi(\mathbf{x})\pi_\theta(\mathbf{y} \mid \mathbf{x}) }{\exp(\beta r(\mathbf{x},\mathbf{y}))}
\right) 
\cdot \nabla_\theta \log \pi_\theta(\mathbf{y} \mid \mathbf{x}) 
\right]
\end{split}
$}
\end{equation}

Next, consider the trajectory balance objective used in GFlowNets learning~\citep{bengio2023gflownet,lee2024learning,bartoldson2025trajectory}, defined as:
\begin{align}
\mathcal{L}(\mathbf{y}, \mathbf{x}; \theta) = 
\left( 
\log \frac{Z_\phi(\mathbf{x}) \, \pi_\theta(\mathbf{y} \mid \mathbf{x})}
{\exp(\beta r(\mathbf{x},\mathbf{y}))} 
\right)^2 .
\end{align}

Taking the gradient of this objective with respect to $\theta$ yields:
\begin{equation}
\begin{split}
\nabla_\theta \mathcal{L}(\theta) = 
2 \cdot \mathbb{E}_{\mathbf{y} \sim \pi_\theta(\cdot \mid \mathbf{x})} 
\left[
\left( 
\log \frac{Z_\phi(\mathbf{x}) \cdot \pi_\theta(\mathbf{y} \mid \mathbf{x})}
{\exp(\beta r(\mathbf{x},\mathbf{y}))}
\right) \cdot \nabla_\theta \log \pi_\theta(\mathbf{y} \mid \mathbf{x})
\right]
\end{split}
\end{equation}

Thus, minimizing the KL divergence is equivalent (up to a constant) to minimizing the trajectory balance loss, confirming Proposition~\ref{prop:kl_gflownet_equivalence}.

\section{Theoretical Analysis}\label{sec:theoretical_analysis}
We conduct an interpretation of FlowRL that clarifies the role of each component in the objective.
\begin{proposition}\label{prop:kl_flowrl_equivalence}
Minimizing the KL divergence in Eq.~\ref{eq:gfn} is equivalent (in terms of gradients) to jointly maximizing reward and policy entropy:
\begin{align}
\max_\theta \; \mathbb{E}_{\mathbf{y} \sim \pi_\theta} \left[
\underbrace{\beta \, r(\mathbf{x}, \mathbf{y})}_{\text{reward}} - \log Z_\phi(\mathbf{x})
+ \log \pi_{\mathrm{ref}}(\mathbf{y}|\mathbf{x}) 
\right] 
+ \underbrace{\mathcal{H}(\pi_\theta)}_{\text{entropy}}.
\label{eq:kl_flowrl}
\end{align}
\end{proposition}

\begin{remark}[\textit{FlowRL beyond reward maximization}]
Proposition~\ref{prop:kl_flowrl_equivalence} reveals that FlowRL can be interpreted as jointly maximizing expected reward and policy entropy. This shift encourages the policy to explore a broader set of high-quality solutions, enabling more diverse and generalizable behaviors on reasoning tasks. Our interpretation also aligns with prior work that views GFlowNets training as a form of maximum entropy RL~\citep{mohammadpour2024maximum,deleu2024discrete}.
\end{remark}
The proof of Proposition~\ref{prop:kl_flowrl_equivalence} is provided as below.

Recall from Eq.~\ref{eq:kl_tb} and Eq.~\ref{eq:gfn} that the FlowRL objective is sourced from the minimization of a KL divergence:
\begin{equation}
\resizebox{\linewidth}{!}{$
\begin{split}
D_{\mathrm{KL}} \left(
\pi_\theta(\mathbf{y} \mid \mathbf{x}) \;\Vert\; 
\frac{
\exp(\beta \; r(\mathbf{x}, \mathbf{y})) \cdot \pi_{\mathrm{ref}}(\mathbf{y} \mid \mathbf{x})
}{
Z_\phi(\mathbf{x})
}
\right) 
= \int \pi_\theta(\mathbf{y} \mid \mathbf{x}) \log 
\left[
\frac{
Z_\phi(\mathbf{x})\pi_\theta(\mathbf{y} \mid \mathbf{x})
}{
\exp\left(\beta \; r(\mathbf{x}, \mathbf{y}) \right) \cdot 
\pi_{\mathrm{ref}}(\mathbf{y} \mid \mathbf{x})
}
\right] d\mathbf{y}
\end{split}$}
\end{equation}

Rearranging the terms, we obtain:
\begin{equation}
\resizebox{0.9\linewidth}{!}{$
\begin{split}
&\argmin_\theta D_{\mathrm{KL}} \left(
\pi_\theta(\mathbf{y} \mid \mathbf{x}) \;\Vert\; 
\frac{
\exp\left(\beta \; r(\mathbf{x}, \mathbf{y}) \right) \cdot \pi_{\mathrm{ref}}(\mathbf{y} \mid \mathbf{x})
}{
Z_\phi(\mathbf{x})
}
\right) \\
&= \argmin_\theta \int \pi_\theta(\mathbf{y} \mid \mathbf{x}) \log 
\left[
\frac{
Z_\phi(\mathbf{x})\pi_\theta(\mathbf{y} \mid \mathbf{x})
}{
\exp\left(\beta \; r(\mathbf{x}, \mathbf{y}) \right) \cdot 
\pi_{\mathrm{ref}}(\mathbf{y} \mid \mathbf{x})
}
\right] d\mathbf{y} \\
&= \argmax_\theta \left\{ \mathbb{E}_{\mathbf{y} \sim \pi_\theta(\cdot \mid \mathbf{x})} \log\left[\frac{\exp\left(\beta \; r(\mathbf{x}, \mathbf{y}) \right) \cdot \pi_{\mathrm{ref}}(\mathbf{y} \mid \mathbf{x})}{
Z_\phi(\mathbf{x})}\right]
- \int \pi_\theta(\mathbf{y}\mid\mathbf{x}) \log\pi_\theta(\mathbf{y}\mid\mathbf{x}) d\mathbf{y} \right\}\\
&= \argmax_\theta \left\{\mathbb{E}_{\mathbf{y} \sim \pi_\theta(\cdot \mid \mathbf{x})} \log\left[\frac{\exp\left(\beta \; r(\mathbf{x}, \mathbf{y}) \right) \cdot \pi_{\mathrm{ref}}(\mathbf{y} \mid \mathbf{x})}{
Z_\phi(\mathbf{x})}\right] +  \mathcal{H}(\pi_\theta) \right\}
\end{split}$}
\end{equation}

Finally, we express the FlowRL objective in its compact form:
\begin{equation}
\max_{\theta} \; \mathbb{E}_{\mathbf{y} \sim \pi_\theta(\cdot|\mathbf{x})} \left[
\underbrace{\beta r(\mathbf{x}, \mathbf{y})}_{\text{reward}} -
\underbrace{\log Z_\phi(\mathbf{x})}_{\text{normalization}} +
\underbrace{\log \pi_{\mathrm{ref}}(\mathbf{y}|\mathbf{x})}_{\text{prior alignment}}
\right]
+ \underbrace{\mathcal{H}(\pi_\theta)}_{\text{entropy}}.
\end{equation}

Therefore, minimizing the FlowRL objective can be interpreted as jointly maximizing reward and entropy, while also aligning the policy with a structured prior. The reward term drives task performance, while the normalization term $Z_\phi(\mathbf{x})$ ensures consistency with a properly normalized target distribution. This encourages the policy $\pi_\theta$ to cover the entire reward-weighted distribution rather than collapsing to a few high-reward modes. The reference policy $\pi_{\mathrm{ref}}$ provides inductive bias that regularizes the policy toward desirable structures, and the entropy term $\mathcal{H}(\pi_\theta)$ encourages diversity in sampled solutions. Together, these components promote better generalization of FlowRL.

\section{GFlowNets}\label{sec:gflownets_notation}
We follow the notation of~\citep{madan2023learning,he2025looking} to introduce the fundamentals of GFlowNets. Let $\mathcal{X}$ denote the compositional objects and $R$ be a reward function that assigns non-negative values to each object $x \in \mathcal{X}$. GFlowNets aim to learn a sequential, constructive sampling policy $\pi$ that generates objects $x$ with probabilities proportional to their rewards, i.e.,$ \pi(x) \propto R(x)$. This process can be represented as a directed acyclic graph (DAG) $\mathcal{G} = (\mathcal{S}, \mathcal{A})$, where the vertices $s \in \mathcal{S}$ are referred to as \textit{states}, and the directed edges $(u \rightarrow v) \in \mathcal{A}$ are called \textit{actions}. The generation of an object $x \in \mathcal{X}$ corresponds to a complete trajectory $ \tau = (s_0 \rightarrow \cdots \rightarrow s_n) \in \mathcal{T}$ within the DAG, beginning at the initial state $s_0$ and ending at a terminal state $s_n \in \mathcal{X}$. The state flow $F(s)$ is defined as a non-negative weight assigned to each state $s \in \mathcal{S}$. The forward policy $P_F(s' \mid s)$ specifies the transition probability to a child state $s'$, while the backward policy $P_B(s \mid s')$ specifies the transition probability to a parent state $s$. To this end, detailed balance objective enforces local flow consistency across every edge $(s \rightarrow s') \in \mathcal{A}$:
\begin{align}
\forall (s \rightarrow s') \in \mathcal{A}, \quad F_\theta(s) P_F(s' \mid s; \theta) = F_\theta(s') P_B(s \mid s'; \theta).
\label{eq:db}
\end{align}
To achieve this flow consistency, GFlowNets employ training objectives at different levels of granularity, including detailed balance~\citep{bengio2023gflownet}, trajectory balance~\citep{malkin2022trajectory}, and sub-trajectory balance~\citep{madan2023learning}. Leveraging their diversity-seeking behavior, GFlowNets have been successfully applied across a range of domains, including molecule generation~\citep{cretu2024synflownet}, diffusion fine-tuning~\citep{liu2025nablagfn,zhang2025improving}, and amortized reasoning~\citep{hu2024amortizing,yuflow}. Among various training objective in GFlowNets, trajectory balance maintains flow consistency at the trajectory level,
% by introducing a learnable normalizing constant 
% Trajectory balance achieve flow consistency on trajectory level. This approach introduces a learnable normalizing constant $Z_\theta$, to estaite the initial flow. 
% The training objective over a complete trajectory is 
defined as:
\begin{align}
Z_\theta \prod_{t=1}^n P_F(s_t \mid s_{t-1}; \theta) = R(x) \prod_{t=1}^n P_B(s_{t-1} \mid s_t; \theta).\label{eq:tb}
\end{align}

Furthermore, sub-trajectory balance achieves local balance on arbitrary subpaths $\tau_{i:j} = \{s_i \rightarrow \cdots \rightarrow s_j\}$, offering a more stable and less biased learning signal. We build on trajectory balance to extend our KL-based objective through a gradient-equivalence formulation (Prop.~\ref{prop:kl_gflownet_equivalence}), and further improve it to better support long CoT reasoning in RL.

\begin{table*}[th!]
  \centering
  \resizebox{\textwidth}{!}{ 
  \begin{tabular}{l|llllll|l}
      \toprule[1.25pt]
        \textbf{Models} 
        & \textbf{AIME 2024} 
        & \textbf{AIME 2025}
        & \textbf{AMC 2023}
        & \textbf{MATH-500}
        & \textbf{Minerva}
        & \textbf{Olympiad}
        & \textbf{Avg} \\
      \midrule[1.1pt]
      \rowcolor[rgb]{0.93,0.93,0.93}
      \multicolumn{8}{c}{\texttt{Qwen2.5-7B Base Model}} \\
      \textcolor{gray}{\texttt{Backbone}} & \textcolor{gray}{4.37} & \textcolor{gray}{2.08} & \textcolor{gray}{30.78} & \textcolor{gray}{54.48} & \textcolor{gray}{22.38} & \textcolor{gray}{24.02} & \textcolor{gray}{23.02} \\
      R++ & $10.57_{\textcolor{scoregreen}{+6.20}}$ & $5.10_{\textcolor{scoregreen}{+3.02}}$ & $66.02_{\textcolor{scoregreen}{+35.24}}$ & $54.29_{\textcolor{red}{-0.19}}$ & $24.47_{\textcolor{scoregreen}{+2.09}}$ & $27.30_{\textcolor{scoregreen}{+3.28}}$ & 31.29 \\
      PPO & $9.95_{\textcolor{scoregreen}{+5.58}}$ & $7.34_{\textcolor{scoregreen}{+5.26}}$ & $63.63_{\textcolor{scoregreen}{+32.85}}$ & $57.72_{\textcolor{scoregreen}{+3.24}}$ & $26.22_{\textcolor{scoregreen}{+3.84}}$ & $27.35_{\textcolor{scoregreen}{+3.33}}$ & 32.03 \\
      GRPO & $14.01_{\textcolor{scoregreen}{+9.64}}$ & $10.73_{\textcolor{scoregreen}{+8.65}}$ & $64.10_{\textcolor{scoregreen}{+33.32}}$ & $57.41_{\textcolor{scoregreen}{+2.93}}$ & $23.17_{\textcolor{scoregreen}{+0.79}}$ & $27.11_{\textcolor{scoregreen}{+3.09}}$ & 32.76 \\
      \midrule
      FlowRL & $14.32_{\textcolor{scoregreen}{+9.95}}$ & $10.05_{\textcolor{scoregreen}{+7.97}}$ & $55.08_{\textcolor{scoregreen}{+24.30}}$ & $66.78_{\textcolor{scoregreen}{+12.30}}$ & $31.52_{\textcolor{scoregreen}{+9.14}}$ & $34.60_{\textcolor{scoregreen}{+10.58}}$ & \textbf{35.39} \\
      \bottomrule[1.25pt]
  \end{tabular}}
  \caption{Math reasoning performance (Avg@64) at temperature $=0.6$. Relative improvements are shown as subscripts, with positive gains in \textcolor{scoregreen}{green} and negative changes in \textcolor{red}{red}. FlowRL consistently outperforms all baselines and achieves the best average score under this low-temperature setting.}
  \label{tab:math_results_agv64}
\end{table*}

\begin{table*}[th!]
  \centering
  \resizebox{\textwidth}{!}{ 
  \begin{tabular}{l|llllll|l}
      \toprule[1.25pt]
        \textbf{Models} 
        & \textbf{AIME 2024} 
        & \textbf{AIME 2025}
        & \textbf{AMC 2023}
        & \textbf{MATH-500}
        & \textbf{Minerva}
        & \textbf{Olympiad}
        & \textbf{Avg} \\
      \midrule[1.1pt]
      \rowcolor[rgb]{0.93,0.93,0.93}
      \multicolumn{8}{c}{\texttt{Qwen2.5-7B Base Model}} \\
      \textcolor{gray}{\texttt{Backbone}} & \textcolor{gray}{3.39} & \textcolor{gray}{1.51} & \textcolor{gray}{23.90} & \textcolor{gray}{45.18} & \textcolor{gray}{16.98} & \textcolor{gray}{18.27} & \textcolor{gray}{18.20} \\
      R++ & $10.63_{\textcolor{scoregreen}{+7.24}}$ & $4.63_{\textcolor{scoregreen}{+3.12}}$ & $66.99_{\textcolor{scoregreen}{+43.09}}$ & $54.36_{\textcolor{scoregreen}{+9.18}}$ & $23.89_{\textcolor{scoregreen}{+6.91}}$ & $26.65_{\textcolor{scoregreen}{+8.38}}$ & 31.19 \\
      PPO & $10.52_{\textcolor{scoregreen}{+7.13}}$ & $6.51_{\textcolor{scoregreen}{+5.00}}$ & $63.04_{\textcolor{scoregreen}{+39.14}}$ & $57.46_{\textcolor{scoregreen}{+12.28}}$ & $25.91_{\textcolor{scoregreen}{+8.93}}$ & $27.16_{\textcolor{scoregreen}{+8.89}}$ & 31.77 \\
      GRPO & $12.50_{\textcolor{scoregreen}{+9.11}}$ & $10.10_{\textcolor{scoregreen}{+8.59}}$ & $64.72_{\textcolor{scoregreen}{+40.82}}$ & $57.15_{\textcolor{scoregreen}{+11.97}}$ & $23.28_{\textcolor{scoregreen}{+6.30}}$ & $26.90_{\textcolor{scoregreen}{+8.63}}$ & 32.44 \\
      \midrule
      FlowRL & $14.22_{\textcolor{scoregreen}{+10.83}}$ & $9.58_{\textcolor{scoregreen}{+8.07}}$ & $52.92_{\textcolor{scoregreen}{+29.02}}$ & $66.20_{\textcolor{scoregreen}{+21.02}}$ & $30.32_{\textcolor{scoregreen}{+13.34}}$ & $34.47_{\textcolor{scoregreen}{+16.20}}$ & \textbf{34.62} \\
      \bottomrule[1.25pt]
  \end{tabular}}
  \caption{Math reasoning performance (Avg@64) at temperature $=1.0$. Relative improvements are shown as subscripts, with positive gains in \textcolor{scoregreen}{green}. FlowRL maintains robust performance under higher generation randomness and continues to outperform all baselines on average.}
  \label{tab:main_results_temp1}
\end{table*}

\begin{table*}[th!]
  \centering
  \resizebox{\textwidth}{!}{ 
  \begin{tabular}{l|cccccc|c}
      \toprule[1.25pt]
        \textbf{Models} 
        & \textbf{AIME 2024} 
        & \textbf{AIME 2025}
        & \textbf{AMC 2023}
        & \textbf{MATH-500}
        & \textbf{Minerva}
        & \textbf{Olympiad}
        & \textbf{Avg} \\
      \midrule[1.1pt]
      $\beta=5$  & 13.54 & 10.00 & 56.09 & 58.91 & 20.79 & 28.72 & 31.34 \\
      $\beta=10$ & 14.79 & 10.20 & 59.53 & 64.30 & 25.27 & 32.39 & 34.41 \\
      $\beta=15$ & 15.41 & 10.83 & 54.53 & 66.96 & 31.41 & 34.61 & 35.63 \\
      % $\beta=\frac{1}{20}$ & 9.791 & 5.65 & 51.09 & 63.42 & 27.43 & 32.31 & 31.62 \\
      $\beta=30$ & 15.00 & 10.83 & 50.62 & 69.02 & 30.03 & 35.03 & 35.09 \\
      \bottomrule[1.25pt]
  \end{tabular}
  }
  \caption{Ablation study on the effect of the $\beta$ parameter in FlowRL. We report Avg@16 accuracy across six math reasoning benchmarks for different values of $\beta$.}
  \label{tab:ablation_flowl}
\end{table*}

\section{Extended Related Work and Comparisons}\label{sec:more_related_work}
Recent notable works have addressed similar challenges in large language model reinforcement learning from different perspectives and across various domains. We provide a detailed comparison below to highlight key distinctions and commonalities with existing methods.

\paragraph{Length Normalization.} Dr.~GRPO \citep{liu2025understanding} proposes an unbiased optimization method that improves token efficiency by removing standard normalization terms from the advantage calculation and removing length terms from the loss objective, while focusing primarily on mathematical reasoning improvements. SRPO \citep{zhang2025srpo} addresses length conflicts through a two-stage training approach (math-first, then coding) and history resampling to filter zero-advantage samples. GSPO \citep{zheng2025group} conducts gradient analysis and applies length normalization in the sequence-level importance ratio ($s_i(\theta) = (\frac{\pi_\theta(y_i|x)}{\pi_{\theta_{\text{old}}}(y_i|x)})^{\frac{1}{|y_i|}}$) to avoid unstable training, particularly crucial for MoE model training. FlowRL operates as a trajectory-level flow-balance objective that initially faced gradient explosion issues during long CoT reasoning. To overcome this challenge, FlowRL integrates length normalization ($\frac{1}{|y|} \log \pi_\theta(y | x)$) directly into the trajectory balance formulation, ensuring training stability and enabling effective scaling to extended CoT sequences. Unlike approaches requiring domain-specific training strategies, FlowRL's unified formulation naturally handles variable sequence lengths through principled reward shaping within the flow-balance framework, achieving stable optimization across diverse reasoning tasks.

\paragraph{KL-Related Policy Optimization Methods.} Kimi-K1.5 \citep{team2025kimi} employs on-policy sampling with KL regularization and uses empirical mean of sampled rewards $\bar{r}$ to approximate the normalizing constant $Z$. This objective has a closed form solution that introduces $\log Z$, where $\gamma$ is a parameter controlling the degree of regularization, maintaining the traditional reward maximization framework.
IPO \citep{azar2024general} addresses overfitting in preference-based learning by using identity mapping ($\Psi = I$) to maintain effective KL regularization with deterministic preferences, targeting preference-based alignment problems. 
FlowRL differs by deriving its objective from reverse KL divergence minimization, shifting from reward maximization to reward distribution matching via flow balance. This approach employs a learnable partition function $Z_\phi(x)$ parameterized by a 3-layer MLP and incorporates importance sampling for the entire trajectory balance objective.
This approach provides both theoretical rigor through generative flow networks and practical effectiveness across diverse reasoning tasks without requiring preference data or domain-specific training paradigms.

\section{Implementation of Partition Function $Z_\phi$}\label{sec:partition_function}

We detail the implementation of the partition function $Z_\phi$, covering theoretical foundations and practical aspects.

From the flow perspective: $Z_\phi$ measures the probability flow from the initial state $S_0$. Intuitively, it estimates the denominator---the sum of rewards across all possible paths---enabling conversion to a probability distribution via $\frac{r(\mathbf{x},\mathbf{y})}{Z_\phi(\mathbf{x})}$.

From the implementation perspective: Since the input of $Z_\phi$ corresponds to the initial state, we utilize the prompt representation from the language model. Specifically, we extract the hidden states from the final layer of the language model for all prompt tokens, and compute their mean to obtain a fixed-dimensional representation. This averaged hidden state vector serves as the input feature for computing the scalar partition function value $Z_\phi(\mathbf{x})$.

% \clearpage
\begin{tcolorbox}[colback=gray!5!white,colframe=gray!75!black,title=Diversity Evaluation Prompt,label={box:diversity-eval}]
\small
\textbf{System:} You are evaluating the DIVERSITY of solution approaches for a mathematics competition problem. Focus on detecting even SUBTLE differences in methodology that indicate different problem-solving strategies.

\textbf{PROBLEM:} \\
\{problem\}

\textbf{16 SOLUTION ATTEMPTS:} \\
\{formatted\_responses\}

\textbf{EVALUATION CRITERIA - Rate diversity from 1 to 5:}

\textbf{Score 1 - Minimal Diversity:}
\begin{itemize}
    \item 14+ responses use essentially identical approaches
    \item Same mathematical setup, same variable choices, same solution path
    \item Only trivial differences (arithmetic, notation, wording)
    \item Indicates very low exploration/diversity in the generation process
\end{itemize}

\textbf{Score 2 - Low Diversity:}
\begin{itemize}
    \item 11-13 responses use the same main approach
    \item 1-2 alternative approaches appear but are rare
    \item Minor variations within the dominant method (different substitutions, orderings)
    \item Some exploration but heavily biased toward one strategy
\end{itemize}

\textbf{Score 3 - Moderate Diversity:}
\begin{itemize}
    \item 7-10 responses use the most common approach
    \item 2-3 distinct alternative approaches present
    \item Noticeable variation in problem setup or mathematical techniques
    \item Balanced mix showing reasonable exploration
\end{itemize}

\textbf{Score 4 - High Diversity:}
\begin{itemize}
    \item 4-6 responses use the most common approach
    \item 3-4 distinct solution strategies well-represented
    \item Multiple mathematical techniques and problem framings
    \item Strong evidence of diverse exploration strategies
\end{itemize}

\textbf{Score 5 - Maximum Diversity:}
\begin{itemize}
    \item No single approach dominates ($\leq$3 responses use same method)
    \item 4+ distinctly different solution strategies
    \item Wide variety of mathematical techniques and creative approaches
    \item Excellent exploration and generation diversity
\end{itemize}

\textbf{IMPORTANT:} Focusing on the DIVERSITY of the attempted approaches. Return ONLY a number from 1 to 5.
\end{tcolorbox}

\end{document}